\documentclass[runningheads]{llncs}

\usepackage{eccv}

\usepackage{eccvabbrv}

\usepackage{graphicx}
\usepackage{booktabs}
\usepackage{multirow}
\usepackage{amsmath}
\usepackage{array}
\DeclareMathOperator*{\argmax}{arg\,max}
\DeclareMathOperator*{\argmin}{arg\,min}

\newcommand{\facet}[0]{{\fontfamily{qcr}\selectfont FacET}}
\newcommand{\translator}[0]{{\textit{translator}}}
\newcommand{\translators}[0]{{\textit{translators}}}

\usepackage[accsupp]{axessibility}  %

\usepackage[pagebackref,breaklinks,colorlinks,citecolor=eccvblue]{hyperref}

\usepackage{orcidlink}
\usepackage{color}
\usepackage{enumitem}

\begin{document}

\title{How Video Meetings Change Your Expression}

\author{Sumit Sarin \and
Utkarsh Mall \and
Purva Tendulkar \and
Carl Vondrick}

\authorrunning{S.~Sarin et al.}

\institute{Columbia University\\
\email{\{ss6712,um2171,pt2578,cv2428\}@columbia.edu}\\[0.22cm]
{\href{https://facet.cs.columbia.edu/}{\textbf{facet.cs.columbia.edu}}}
}

\maketitle

\begin{abstract}
 Do our facial expressions change when we speak over video calls? Given two unpaired sets of videos of people, we seek to automatically find spatio-temporal patterns that are distinctive of each set. Existing methods use discriminative approaches and perform post-hoc explainability analysis. 
 Such methods are insufficient as they are unable to provide insights beyond obvious dataset biases, and the explanations are useful only if humans themselves are good at the task. 
 Instead, we tackle the problem through the lens of generative domain translation: our method generates a detailed report of learned, input-dependent spatio-temporal features and the extent to which they vary between the domains. We demonstrate that our method can \textit{discover} behavioral differences between conversing face-to-face (F2F) and on video-calls (VCs). We also show the applicability of our method on discovering differences in presidential communication styles. 
 Additionally, we are able to predict temporal change-points in videos that decouple expressions in an \textit{unsupervised} way, and increase the interpretability and usefulness of our model. 
 Finally, our method, being generative, can be used to transform a video call to appear as if it were recorded in a F2F setting. Experiments and visualizations show our approach is able to discover a range of behaviors, taking a step towards deeper understanding of human behaviors.
\end{abstract}

\section{Introduction}
\label{sec:intro}

\begin{figure}[tb]
  \centering
  \includegraphics[width=1.\linewidth]{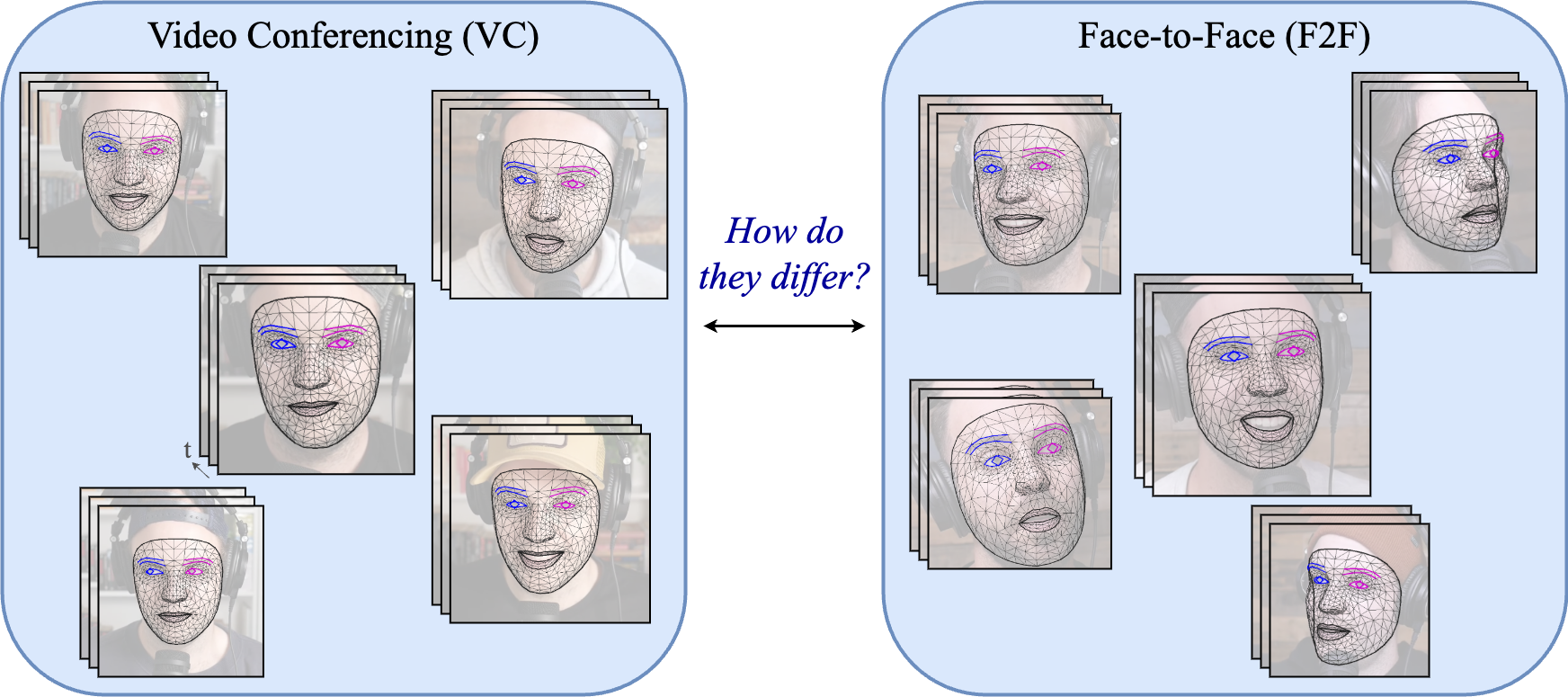}
  \vspace{-10pt}
  \caption{
  \textbf{What is the difference between the two domains?}
    Given two unpaired sets of videos of persons speaking on VC (left) and F2F (right), our goal is to provide interpretable insights on how the motion sequence differs between the domains. See Figs.~\ref{fig:results-faces} \& \ref{fig:main} for a detailed report generated by our approach.
  }
  \label{fig:teaser}
  \vspace{-5pt}
\end{figure}

Have you wondered how your facial expressions change when speaking to someone over a video call (VC) as compared to speaking face-to-face (F2F) (Fig.~\ref{fig:teaser})? Studies have shown that there are significant differences between these two modes of communication -- in terms of overall effectiveness, rapport building, cognitive load, energy consumption, etc. -- which can manifest itself in changed expression patterns~\cite{zhao2023separable, Balters_zoom2023, matz_personality2020, khan_virtual2021, archibaldzoom2019}. Since the recent intervention of COVID-19, our lives drastically changed as VC became the primary source of communication, resulting in unprecedented social phenomena such as \emph{zoom fatigue}~\cite{nesher-22, FAUVILLE2021100119, Bailenson2021Nonverbal, boland_zoomrhythm, Fauville_zoom2021}. It is important to systematically study these changed patterns in order to build tools that can cope with such interventions -- e.g., by evaluating and improving AR/VR technology so that our virtual experiences are as similar to reality as possible. Moreover, studying changes in human behavior, at an individual-level, as well as across a population, can be useful in a variety of disciplines including anthropology, sociology, as well as cognitive and social psychology~\cite{hoehe2020, numata2020, smith2018}.

\begin{figure}[t!]
    \begin{minipage}{0.2\textwidth}
        \centering
        \includegraphics[width=\linewidth]{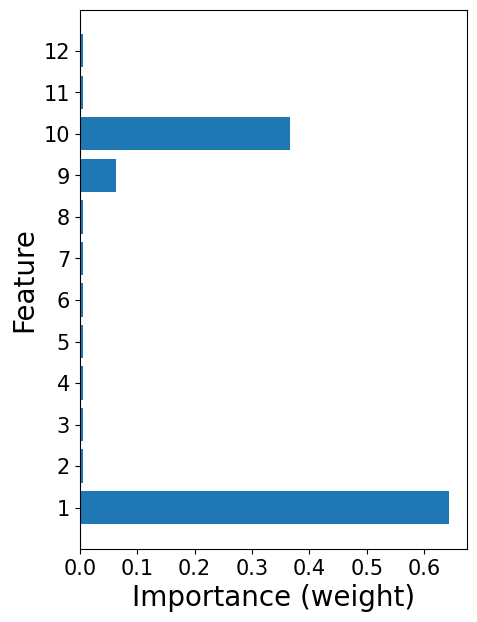}
    \end{minipage}
    \begin{minipage}{0.03\textwidth}
    \end{minipage}
    \begin{minipage}{0.75\textwidth}
  \caption{
    \textbf{Insufficiency of discriminative methods.} 
    We train a simple linear classifier on disentangled facial features to distinguish VC and F2F conversations (refer to~\cref{fig:results-faces} for the meanings of these features). 
    We observe 88\% classification accuracy even when using frames without the temporal information, with `Head Pitch' (\#1) and `Head Tilt' (\#10) being the most dominant features. 
    A post-hoc explainability model cannot explain this discriminative model as it is trained on biased data. 
  }
  \label{fig:bar}
    \end{minipage}
  \vspace{-15pt}
\end{figure}

Building models for understanding facial expression changes in conversations is challenging for two key reasons. 
\emph{First} is the dataset bias. 
Internet videos are an abundant source of natural and diverse data~\cite{geng2023affective}.
However, we found that internet videos of people on VC and F2F are heavily biased.  VCs often have the person looking directly at the camera, while F2F has side-facing views. These are not \textit{inherently} harmful, but a model that simply identifies these is not sufficient to \textit{discover} all the differences. 
Unsurprisingly, we find that it is quite easy to build a classifier that can differentiate the two domains.~\cref{fig:bar} shows the weights of a linear classifier trained on disentangled facial features (see \cref{ssec:bvae}). 
With no temporal information, the classifier attains 88\% accuracy on single frames alone! 
Moreover, these biases are implicit, and apriori unbeknownst to us, thus making it near impossible to remove.

\emph{Second}, our task is not simply that of classification, but rather \emph{discovery of domain-specific differences.} 
We lack prior knowledge about the domains, and we wish to understand them better. 
Being able to predict the domain is perhaps less interesting than being able to understand the subtle spatio-temporal differences of facial expressions across the two domains. 
Much of the current literature focuses on post-hoc explainability of black-boxes~\cite{fong2019understanding, petsiuk2018rise, selvaraju2017grad, shitole2021one, zeiler2014visualizing, zhou2016learning}. 
However, such approaches are unsuitable for our goals because discriminative models (such as~\cref{fig:bar}) trained on biased data only pick up these biases. 
Additionally, the explanations are useful only if humans themselves are good at the task.

We present~\facet, a general-purpose framework to discover interpretable, spatio-temporal trends between two domains, that works even in the presence of dominant biases. 
In contrast to discriminative methods,~\facet~employs a generative approach to discover the differences.
To study the aforementioned task of VC and F2F conversations, we first collect a large video dataset (240 hours) of such conversations, called the ZoomIn dataset. In addition, we also collect a second dataset of speaking styles of US presidents.
We apply~\facet~on these datasets, and reveal novel insights -- e.g., we observe that speakers tend to laugh \textit{bigger} in F2F as compared to VC. We also observe that President Trump raises his eyebrows more while listening compared to President Obama.
Moreover,~\facet~has applications beyond revealing these insights -- we show that~\facet~can be used to perform domain transfer -- e.g., we can modify VC videos to look more like a F2F conversation (effectively "de-zoomifying" a video).

\section{Related Work}
\label{sec:related}

We discuss the landscape of interpretability in computer vision by describing post-hoc explanations, interpretable architectures, as well as disentangled representations. We then discuss efforts in human facial expression understanding.\\

\noindent\textbf{Interpretability through post-hoc explanations. } 
Explainable AI (XAI)~\cite{Gunning_Aha_2019} methods seek to explain model decisions through visualizations~\cite{fong2019understanding, petsiuk2018rise, selvaraju2017grad, shitole2021one, zeiler2014visualizing, zhou2016learning}, counterfactual
explanations~\cite{goyal2019counterfactual, vandenhende2022making, wang2020scout}, feature importance~\cite{ribeiro2016should} and sample importance~\cite{koh2017understanding, yeh2018representer, Tsai2023SampleBE, Sui2021RepresenterPS, Pruthi2020EstimatingTD, Silva2020CrossLossIF, Guo2020FastIFSI} methods.
These explanations are important to ensure algorithmic fairness~\cite{pan2021explaining}, identify potential bias in the training data~\cite{pradhan2022interpretable, meng2022interpretability, alelyani2021detection}, and to ensure that the algorithms perform as expected~\cite{gilpin2018explaining, kim2022hive, selvaraju2020squinting, das2017human}. 
However, post-hoc explanations are only meaningful if humans are good at solving these tasks, and therefore check the explanation against their understanding. 
Also they can only explain what the model has learned, which is problematic if the data is biased.
Moreover, they are more useful for explaining model decisions than data patterns, which is our focus. \\

\noindent\textbf{Interpretability by design. } 
A number of works have defined interpretability through a set of linear transforms between classes~\cite{brendel2019approximating, bohle2021convolutional, bohle2022b, chen2019looks, donnelly2022deformable, pmlr-v119-koh20a}.
However these methods are limited to discriminative tasks where classes are distinguishable by humans.
Generalized Additive Model~\cite{hastie2017generalized} and its variants ~\cite{Lou2013AccurateIM, dubey2022scalable, radenovic2022neural, chang2021node, agarwal2021neural} are another class of methods which non-linearly transform input features separately. 
However, these methods suffer from complex training procedures and combinatorial explosion.
Our method also falls in this category of interpretability-by-design, and leverages linear transformations in a generative translation setting. \\

\noindent \textbf{Learning Disentangled Representations. } Learning factorized representations to capture independent variations in data can correspond to human-defined concepts and lead to interpretable models~\cite{Burgess2018UnderstandingDI}. Learning disentangled representations using supervised data (apriori knowledge of the nature of generative factors) has been explored by many methods~\cite{Zhu2014MultiViewPA, Reed2014LearningTD, Whitney2016UnderstandingVC, Cheung2014DiscoveringHF}. However, for practical applications where there is no supervision available for discovering the generative factors underlying the data, purely unsupervised approaches that use GANs~\cite{Lin2019InfoGANCRDG, Jeon2018IBGANDR, Chen2016InfoGANIR, Ramesh2018ASR, dalva2022vecgan, dalva2023face} and VAEs ~\cite{higgins2017betavae, Kim2018DisentanglingBF, Chen2018IsolatingSO, Jeong2019LearningDA, Kumar2017VariationalIO} for learning disentangled representations are more useful. We use $\beta$-VAE~\cite{higgins2017betavae} which is one such popular unsupervised method.\\

\noindent\textbf{Facial Expressions Understanding. } 
Existing literature on human facial expression understanding is limited to classifying emotions into known categories~\cite{yesubig5, KNYAZEV20081093, assessingbig52020, budenbender2023, StahelskiFacial2021, SnoekTesting2023, Straulino_emotion2023}, or their combinations~\cite{ducompound2014}.
Several works have also studied person-specific speaking styles~\cite{Minetaki_facial2023, yesubig5} which link facial expression changes to personality traits. 
However, the continuous nature of facial expressions is not accurately modeled through such discrete representations.
Recent works have also studied how a person's facial expressions change in response to the person they are interacting with~\cite{jonell2020let, geng2023affective, ng2023text2listen, ng2022learning2listen}.
Yet, none of these works provide interpretable insights regarding inter or intra-subject patterns under varied conditions.

\section{Method} \label{sec:method}
A conventional approach to finding interpretable differences between two domains is to learn a discriminative model to distinguish and analyze them using post-hoc tools such as GradCAM~\cite{selvaraju2017grad}.
However, discriminative models tend to focus on the first few significant modes of differences. 
Consequently, we cannot find all the important differences.
This problem is further exacerbated if the dataset is biased, and significant differences appear only due to such biases. 
We posit that a generative model that can transform one domain to another should capture all possible modes of differences. 
Such a generative model can then be leveraged to analyse the different modes and their effect on the two domains.

We present~\facet~ (\textbf{Fac}ial \textbf{E}xplanations through \textbf{T}ranslations). 
Given spatio-temporal facial keypoint data from two domains $X$ and $Y$,~\facet~learns a translation function $G_{XY}: X \rightarrow Y$, that can convert samples from domain $X$ to appear as if they were from domain $Y$.
Our key insight is to \emph{learn piecewise shift and scale transformations}~\cite{brendel2019approximating, bohle2021convolutional, bohle2022b, chen2019looks, donnelly2022deformable, pmlr-v119-koh20a} between $X$ and $Y$ which results in $G_{XY}$ being interpretable, allowing us to explain the modes of differences between the two domains. 
To train~\facet, we first learn a per-frame spatial representation to disentangle facial features (\cref{ssec:bvae}). 
Then, we train an interpretable $G_{XY}$ on these features capturing spatio-temporal differences between the domains (\cref{ssec:temporal}). 
We use the trained $G_{XY}$ to generate detailed reports (\cref{ssec:interpretability}) highlighting key differences between the two domains.
Finally, we showcase ~\facet's ability to perform domain translation, e.g., by ``de-zoomifying'' a VC video (\cref{ssec:dezoom}). 

\subsection{Disentangling Spatial Features}\label{ssec:bvae}
The first step towards an interpretable translation model is learning a disentangled spatial representation. 
We found that disentangled spatio-temporal features are difficult to interpret for humans. 
Thus, we first learn to disentangle spatial features only.
$\beta$-VAEs~\cite{higgins2017betavae} have been found to generate disentangled representations that are useful in discovering patterns in a variety of domains~\cite{burgess2018understanding, li2023betavae, garcia2022towards, pastrana2022disentangling, Higgins_2021}.
We train a $\beta$-VAE on frame-level facial keypoints of our datasets. 
We obtain a spatial feature $z\in \mathbb{R}^l$ for a data point $d$, where $d$ is a single frame in the clips of our dataset $X\cup Y$.
Here $l$ is the dimension of latent space.
This can be achieved by learning an encoder ($q$) and decoder ($p$) by minimizing 
the $\beta$-VAE objective,

\begin{equation}
      \mathcal{L}(X\cup Y) = -\mathbb{E}_{q(z|d)}[\log p(d|z)] + \beta D_{KL}(q(z|d)||p(z))
  \label{eq:betavae}
\end{equation}

The first term (marginal log-likelihood) optimizes for reconstruction quality, while the second term (KL-divergence) enforces disentanglement in the latent space $z$. 
Minimizing the KL divergence between the latent vectors and a unit Gaussian prior ($p(z)=\mathcal{N}(0, \mathbf{I})$), results in disentangled latent vectors. 
Higher $\beta$ values result in better disentanglement at the cost of reconstruction.

\begin{figure}[t!]
  \centering
  \includegraphics[width=1.\linewidth]{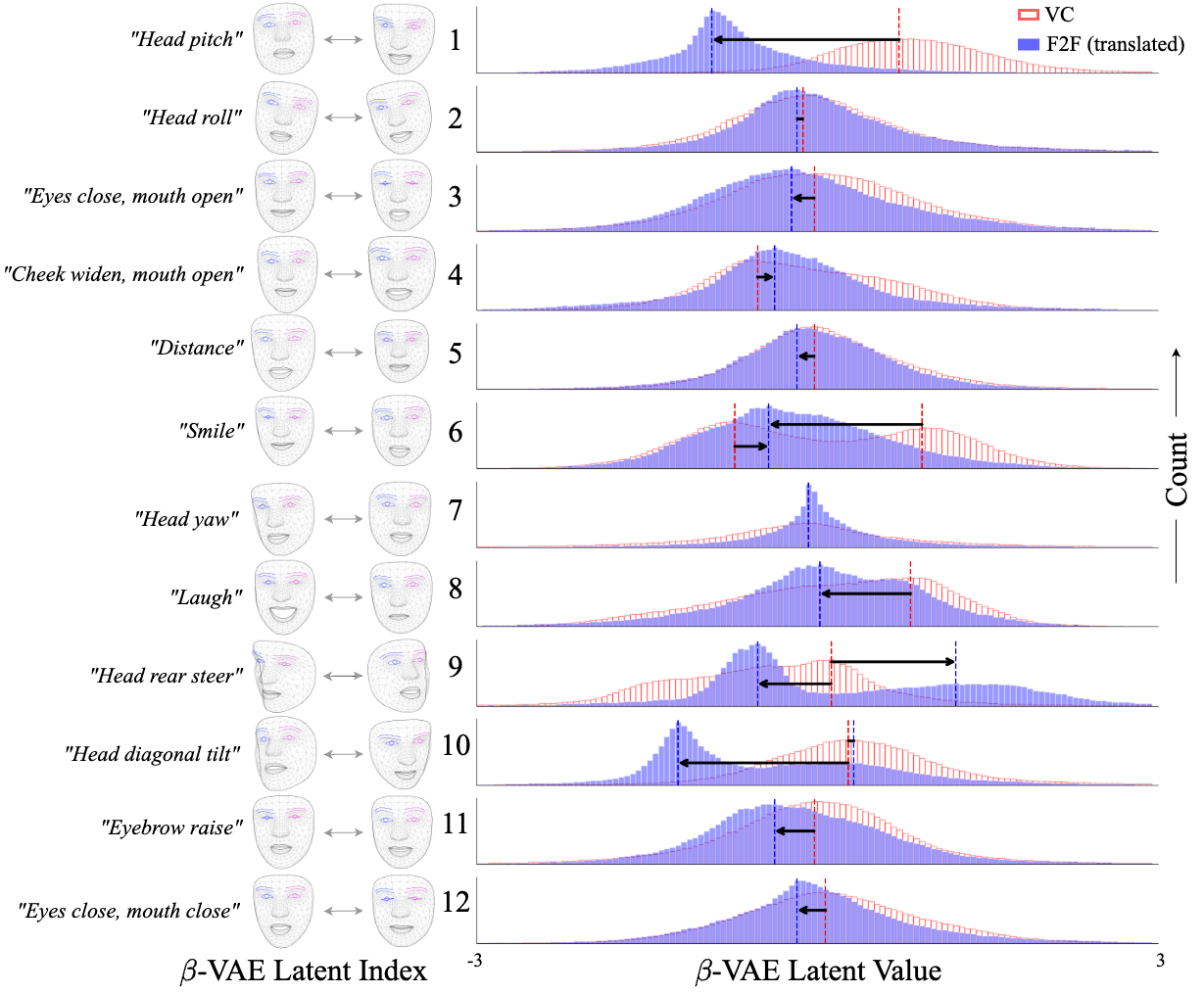}
  \vspace{-15pt}
  \caption{
    \textbf{Explanation of Disentangled Latents.} We vary each dimension of the 12-dimensional latent obtained through $\beta$-VAE encoding by keeping other dimensions fixed (rows). \textbf{Left:} The faces corresponding to the extreme values of the perturbed latent, along with a description of the dominant change. \textbf{Right:} Dataset-level statistics for the desired latent, while the dotted line shows the modes. The direction and length of the arrows show the extent to which different latents change across domains on average. 
    Please refer to the supplementary for videos visualizing the latents.
  }
  \label{fig:results-faces}
  \vspace{-15pt}
\end{figure}

\subsection{Translation Function}\label{ssec:temporal} 
We learn the translation function $G_{XY}$ in the $\beta$-VAE's spatial latent space $z$. 
We aim to learn a $G_{XY}$ that can potentially capture all differences between the two domains.
To achieve this, we employ an adversarial discriminator objective. 
The goal of the discriminator is to keep finding differences between real and translated (fake) samples so that $G_{XY}$ keeps improving.
We denote the discriminator differentiating real and fake samples from domain $Y$ as $D_{Y}$.
For a clip $x\in X$, we denote its latent encoding as $z^x$ (similarly, $z^y$ for a sample $y\in Y$). 
By encoding individual frames of $x = \{x_1\cdots x_t\}$ we can obtain the latent encodings $z^x = \{z^x_1,\cdots z^x_t\}$, where $z^x_i = q(x_i)$.
The adversarial loss can be written as,
\begin{equation}
  \mathcal{L}_{adv}(X, Y) = \mathbb{E}_{y\in Y}[\log D_Y(z^y)] + \mathbb{E}_{x\in X}[\log(1-D_Y(G_{XY}(z^x)))]
  \label{eq:adv}
\end{equation}

Similar to other adversarial objectives~\cite{goodfellow-14, zhu2017unpaired}, we optimize this loss function by alternatively training the translation function $G_{XY}$ and the discriminator $D_Y$.
The translation function $G_{XY}$ can be found using the following optimization,
\begin{equation}
  G_{XY}^* = \argmin_{G_{XY}} \max_D L_{adv}(X,Y)
  \label{eq:opt}
\end{equation}

\noindent \textbf{Translation Function Architecture.} 
If we do not constrain $G_{XY}$ -- for example, by using an MLP for $G_{XY}$  -- it would not be interpretable. Moreover, if the capacity of the network is high enough, the function could learn arbitrarily complex mapping between domains. For example, it could memorize a one-to-one mapping between examples from two domains, which is not a useful translation.

To tackle both these problems we need to constrain $G_{XY}$.
Our key insight is that instead of directly learning to translate from domain $X$ to $Y$, $G_{XY}$ 
first predicts a~\translator~function $f$, that when applied to a sample $z^x$ results in the translated output ${z^{y}}' = f(z^x)$.
The advantage of predicting such a~\translator~is that its parameters are more interpretable than a black box model.
For example, as we will see in Fig.~\ref{fig:timestamp}, our model $G_{XY}$ finds consistent \emph{translators} for clips with the same expressions such as ``smiling'' or ``listening''.
We also parameterize the~\translator~$f$ to be a \emph{shift and scale operation}, using a multiplicative factor $\omega \in \mathbb{R}^l$ and an additive factor $\phi \in \mathbb{R}^l$. 
Therefore the~\translator~$f(x) = \omega \odot z + \phi$.

One issue with the above formulation is that predicting the same~\translator~$f$ for the entire clip can be too restrictive. 
The expressions and modes of conversation change within a clip. 
For example, a listener might start speaking or a speaker might exclaim in the middle of a clip.
Therefore, instead of predicting a single~\translator~$f$ for the entire clip, we predict different~\translators~for different chunks. 
This also makes the model more robust as it is forced to learn similar~\translators~for similar segments.
We achieve this by breaking down $G_{XY}$ into two sub-modules: $G_{t}$ and $G_{f}$. $G_{t}$ first predicts $c-1$ temporal change-points $\{\tau_1, \cdots \tau_{c-1}\}$ partition the clip into $c$ chunks. $G_{f}$ then predicts a~\translator~for each chunk -- e.g., for the $k^{th}$ chunk, $G_{f}$ predicts the~\translator~$f_k$ ($\omega_k, \phi_k$).

A key challenge here is that we do not have supervision to learn how to partition a clip into semantically meaningful chunks. In fact, discovering such meaningful chunks from the data is one of our goals. Therefore, we learn both the functions $G_{f}$ and $G_{t}$ together in an end-to-end fashion.
Note that partitioning a clip using values from $G_{t}$ and passing each chunk through $G_{f}$ is a non-differentiable operation. To circumvent this, we approximate this partitioning operation with a continuous alternative. For each chunk $k$, we learn smooth weights over time $\bar{w}_k \in [0, 1]^t$, where $t$ is the total clip length. If $T$ is the vector of time indexes $[1, 2, \cdots t]$  for a clip, then we define $\bar{w}_k$ as:
\begin{equation}
\begin{aligned}
w_k = 
\begin{cases}
    \sigma(\tau_k-T, Q) & k = 1 \\
    \min(\sigma (T - \tau_{k-1}, Q), \sigma (\tau_k-T, Q)) & k \in [2, c-1]\\
    \sigma (T - \tau_{k-1}, Q) & k=c
\end{cases} 
\qquad
& \bar{w}_k = \frac{w_k}{\sum_{k=1}^{c}w_k} 
\end{aligned}
\end{equation}

\begin{figure}[t!]
  \centering
  \includegraphics[width=1.\linewidth]{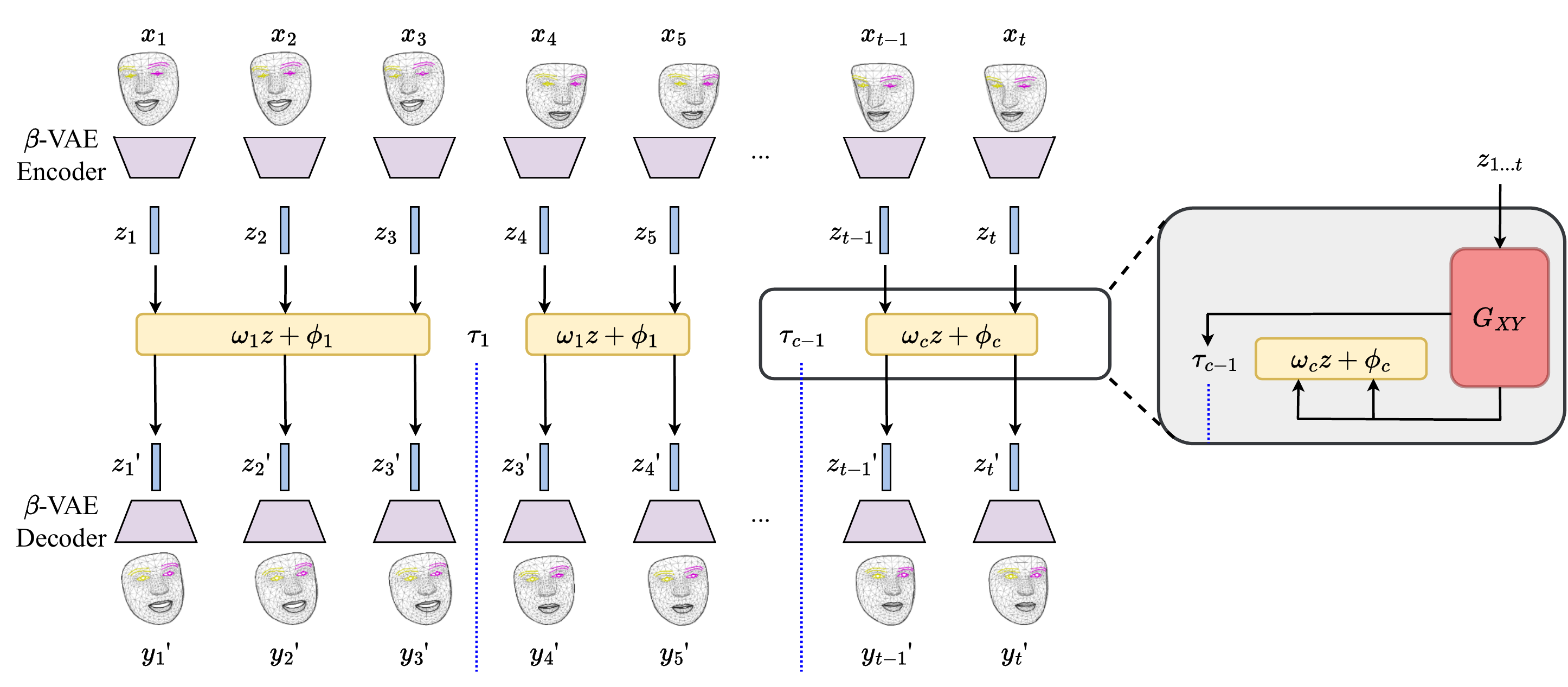}
  \vspace{-15pt}
  \caption{
    \textbf{Method Overview.} Given a sequence of facial keypoints $x$, we use a pre-trained $\beta$-VAE encoder to obtain latents $z$. We then train a translation function $G_{XY}$ that takes as input the latents $z$ to produce a \translator ($\omega$, $\phi$, $\tau$). This~\translator~when applied to $z$ generates the transformed latent $z'$, which can be decoded using the $\beta$-VAE decoder to obtain the transformed facial keypoints $y'$ belonging in the new domain. 
  }
  \label{fig:method}
  \vspace{-15pt}
\end{figure}

Here $\sigma$ is the sigmoid function, and $Q$ is the temperature. Essentially, ${w}_k$ is a continuous alternative of a non-differentiable rectangular pulse function. $\bar{w}_k$ normalizes ${w}_k$, so that they sum up to 1 at every time step. 
With $\bar{w}_k$, a differentiable~\translator~ predictor can be written as $G_f(z^x \odot \bar{w}_k)$ for chunk $k$.

To summarize, our model minimizes the optimization problem defined in Eq.~\ref{eq:opt} to predict the parameters of the two networks $G_t$ and $G_f$, which together represent $G_{XY}$. An overview of this model is illustrated in \cref{fig:method}

\vspace{-5pt}
\subsection{Interpretability}\label{ssec:interpretability}
Our translation function is interpretable by design. 
First, the model can partition a clip into semantically uniform chunks without supervision.
Second, predicting a~\translator~instead of predicting the translation function $G_{XY}$, allows us to understand how a particular expression or action changes between domains. For e.g., we can answer how ``listening'' changes between VC and F2F conversations.
Finally, modeling a~\translator~as a shift and scale operation, allows us to factor the translations into individual $\beta$-VAE latents. For e.g., we can answer whether people ``smile more or less'' when ``listening'' in VC versus F2F conversations.

Note that formulating the~\translator~to be a shift and scale operation is a modeling choice. 
Depending on the task and the sample representation, we can adjust the parametrization. 
Also note that our framework closely resembles CycleGAN \cite{zhu2017unpaired} with its adversarial objective, albeit without the cycle consistency loss. 
Cycle consistency is used to over-constrain the model to prevent it from reaching trivial solutions.
Our model, being over-constrained by design, cannot lead to such a trivial solution. 
Moreover, even with cycle consistency, a model with enough capacity can memorize a one-to-one mapping between examples from the two domains~\cite{zhu2017unpaired, NEURIPS2022_958b6310}. 
This is not a problem in cases where we as humans have prior knowledge and can qualitatively show that the outputs change in accordance with our prior knowledge. 
However, we do not have prior knowledge of what makes VC and F2F different. 
Our constraints prevent the model from learning such trivial solutions \cite{shen2020one}.

\vspace{-5pt}
\subsection{Domain Transfer}\label{ssec:dezoom}
Besides systematically discovering differences across domains,~\facet~being generative in nature, can also transform inputs ($x \rightarrow y'$) from domain $X$ to $Y$. To visualize the results in pixel-space, we leverage a video-to-video model, vid2vid~\cite{wang2018vid2vid}, that learns to map keypoints to pixels using our dataset. Compared to existing methods that simply correct for  
gaze to make VC more like F2F~\cite{KusterGaze2012}, we expect that style-transfer using~\facet~will 
lead to more faithful transfer that encompasses subtle expression changes as well.

\section{Experiments}\label{sec:experiments}

In this section, we show how our model can discover differences between domains.
We first describe the two datasets that we collected to study this task (\cref{ssec:dataset}). 
Then we evaluate~\facet~quantitatively (\cref{ssec:quant}) on these datasets, and finally discuss the interpretable outcomes of using it (\cref{ssec:quali}). Please refer to the supplementary for implementation details.
\subsection{Datasets}\label{ssec:dataset}
To study the differences between VC and F2F, we collect the ZoomIn dataset from YouTube, containing 240 hours of conversations in both domains (See~\cref{fig:teaser} for examples).
Additionally, to study differences in communication styles across subjects, we collect a dataset of 20 hours of individual speaking styles of US presidents Obama and Trump.
Each clip in our datasets consists of detections on contiguous 176 frames from both the participants. For more information on these datasets please refer to the supplementary.

\subsection{Quantitative Results}\label{ssec:quant}

\subsubsection{Metrics.}
Evaluating discovered differences is challenging in the absence of a paired dataset, since we cannot evaluate what will be discovered beforehand.
Thus, our metric is the translation function~$G_{XY}$'s ability to fool the discriminator $D$, while being interpretable. This is measured using discriminator accuracy. We notice that this value tends to stabilize after a period of training, and hence we provide the average across the last 100 epochs to ensure robustness. Please refer to the supplementary for more discussion on metrics.\\

\begin{table}[ht!]
  \caption{Performance of our method by its ability to fool the discriminator whilst being interpretable (lower is better, 50\% is optimal).}
  \label{tab:ablation}

  \centering
  \begin{tabular}{@{}llr@{\hspace{0.3cm}}ccc@{\hspace{0.3cm}}ccc@{}}
    \toprule
    \multicolumn{3}{c}{\textbf{Model Type}}  & \multicolumn{6}{c}{\textbf{Discriminator Acc. (\%) $\downarrow$}} \\
    & &  & \multicolumn{3}{c}{ZoomIn} & \multicolumn{3}{c}{Presidents} \\
    \cmidrule(r{1em}){1-3}
    \cmidrule(lr{2em}){4-6}
    \cmidrule(lr{0.0em}){7-9}
        $G_f$ & $G_t$ & c & Avg & \hspace{0.cm}F2F$\rightarrow$VC & \hspace{0.cm}VC$\rightarrow$F2F  & Avg & \hspace{0.1cm}O$\rightarrow$T & \hspace{0.1cm}T$\rightarrow$O  \\
    \midrule
    \multirow{5}{0.12\linewidth}{Fixed translator set} & No Partitions & 1 & 87.58 & 87.92 & 87.06 & 81.40 & 71.33 & 92.19 \\
    \cmidrule{2-9}
    & Fixed-size Chunks & 2 & 93.95 & 93.23 & 94.70 & 90.95 & 86.90 & 95.21 \\
    & Fixed-size Chunks & 7 & 97.78 & 98.54 & 97.02 & 96.64 & 94.81 & 98.63 \\
    \cmidrule{2-9}
    & Var. Chunks & 2 & 92.26 & 91.91 & 92.62 & 83.80 & 75.14 & 92.63 \\
    & Var. Chunks & 7 & 97.67 & 97.18 & 98.15 & 97.17 & 95.54 & 98.76 \\
    \midrule
    \multirow{5}{0.12\linewidth}{Predicted translator} & No-partitions & 1 & 78.54 & 79.71 & 77.71 & 79.25 & 69.88 & 88.97 \\
    \cmidrule{2-9}
     &Fixed-size Chunks & 2 & 82.99 & 82.10 & 84.12 & 88.67 & 87.63 & 89.42 \\
     &Fixed-size Chunks & 7 & 81.84 & 81.06 & 82.95 & 89.86 & 89.73 & 89.66 \\
    \cmidrule{2-9}
    & {Var. Chunks (\textbf{\facet})} & 2 & 73.28 & 73.33 & \textbf{73.50} & 79.35 & 70.31 & \textbf{88.70} \\
    & {Var. Chunks (\textbf{\facet})} & 7 & \textbf{73.16} & \textbf{72.65} & 73.92 & \textbf{78.14} & \textbf{67.50} & 89.06 \\
    \bottomrule
  \end{tabular}
\end{table}

\noindent \textbf{Baselines.}
To the best of our knowledge, ours is the first work that tackles this problem, so there are no comparable baselines.
Thus, we validate our design choices through the following interpretable ablations of our model:
\begin{itemize}[nosep,leftmargin=*]
\item \textbf{No Partitions:} a model where we predict a single~\translator~for the entire duration of the clip i.e. $c=1$. Note that this model does not need $G_t$.
\item\textbf{Fixed-size Chunks:} a model with uniform, equal-sized chunks. This model also does not use $G_t$. However $G_f$ predicts different~\translator s for each chunk.
\item\textbf{Fixed translator-set:} instead of \emph{predicting} the parameters of the \translator, our model \emph{retrieves} parameters from a set of $p$ options. We also learn this fixed-size set of parameters $P$.
$G_f$ then becomes a $p$-way classifier that takes as input a chunk and predicts one of the $p$ classes using the softmax function. 
Since the parameters are $\omega, \phi \in \mathbb{R}^l$, $P\in \mathbb{R}^{p\times 2l}$ is a matrix.
We jointly optimize both the classifier and the parameters $P$. We use $p=32$ for our experiments.
\end{itemize}

\noindent \textbf{Is \facet~better at translation than the baselines?} \cref{tab:ablation} shows the performance of \facet~and baselines using discriminator accuracy. 
We learn translation functions in both directions between the domains.

\facet~performs better translations than the baselines and consequently results in the lowest discriminator accuracy.
With uniform equal-sized chunks, $G_f$ cannot accurately predict good~\translator s.
The facial expressions may change in the middle of the chunk, but this model would apply the same translation function even if the expression changes, resulting in poor translations.
Similarly, if we do not partition the clip, the translations are also of poor quality.
Swapping the model predicting the~\translator~with a model selecting~\translator s from a set also results in poor translations.
We argue that a fixed set of~\translators~is too restrictive and cannot model all possible expressions even with a larger $p$.
For~\facet~increasing the number of chunks from 2 to 7 does not affect the performance by a lot. 
In the presidential data, since the utterances are often monotonous and in a ``speech''-like non-dyadic manner, more chunks are not beneficial to model performance.
We believe, that due to the nature of conversations in podcasts, the expressions often do not change more than once in a 7-second clip.
Therefore, we use \facet~with 2 chunks for qualitative analysis. 

\subsection{Qualitative Results}\label{ssec:quali}

Now that we have demonstrated that \facet~can faithfully translate samples, we look at what we can discover on our data.

\noindent \textbf{How do F2F conversations differ from VC?} 
We answer this question by using \facet.
We then look at the distribution of latent codes for F2F and translated VC obtained from \facet.
\cref{fig:results-faces} shows the differences between VC and F2F along the 12 latent codes.
Some of these differences are quite apparent, for example, people look down with respect to the camera in VC.
This is because the cameras are generally above the screen where the person is looking during a VC.
This is evident in the distribution of latent code \#1, where \facet~rotates the head upwards when translating from VC to F2F.

\begin{figure}[t!]
  \centering
  \includegraphics[width=0.99\linewidth]{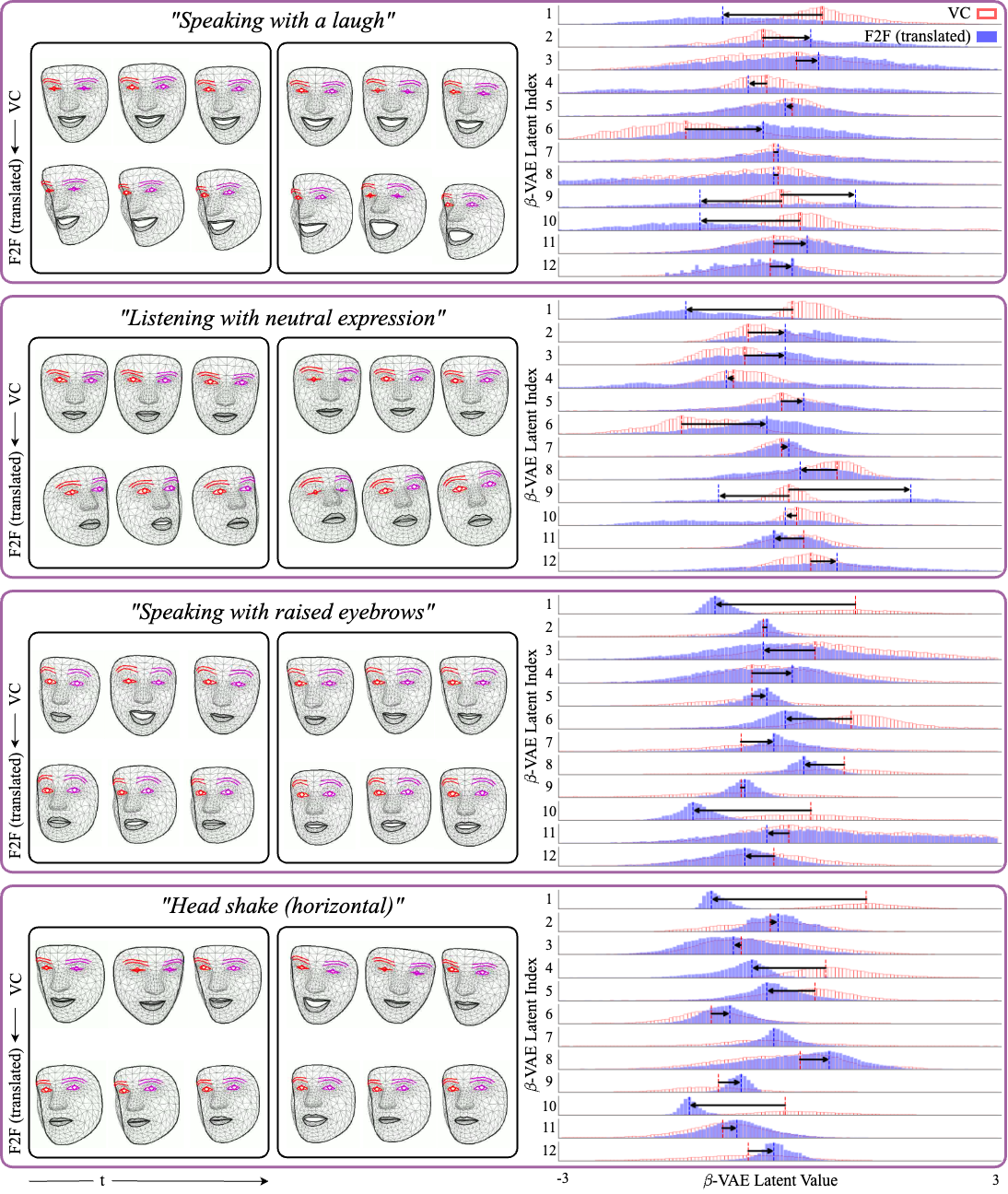}
  \vspace{-5pt}
  \caption{
    \textbf{Results.} We showcase our key findings of domain differences through a detailed report. \textbf{Left:} Each row corresponds to two different examples belonging to a specific~\translator~cluster (e.g.,\textit{"Speaking with a smile"}). The top row shows a video recorded in VC while the bottom row shows the transformed video as if it were F2F. \textbf{Right:}\facet~generates a report for each corresponding cluster showing how each $\beta$-VAE latent varies across the domains. Refer to Fig.~\ref{fig:results-faces} to interpret each latent index. See Sec.~\ref{ssec:quali} for detailed explanations. Please refer to supplementary for full videos.
  }
  \label{fig:main}
  \vspace{-15pt}
\end{figure}

Similarly, we observe big changes in head tilt (\#10) and head steer (\#9), because in VC the face often always stays towards the screen, whereas the orientation of the head can vary a lot in F2F.
We also notice that the head tilt (\#10) and head steer (\#9) distributions become bimodal after translation (from unimodal). 
This can again be attributed to the fact that, during a VC, we only look at screens, however in a 2-person F2F conversation, we will see two modes with peaks at the orientation where the subjects are looking at one another.

Some other observations are not quite easily discernible but are evident from our method.
We see that the the smile latent distribution (\#6) translates from a bimodal distribution in VC to a unimodal distribution with a peak in the center in F2F. 
Similarly, the eyebrow raise latent (\#11) becomes very muted in an F2F conversation.
We conjecture that people tend to emote more during a VC as smaller reactions are harder to pick in a VC~\cite{faith2023zoomgesture}. 
On the contrary, we notice that people laugh less during a VC. 
While it is easy to emote silently on Zoom, laughing is harder because VC systems usually allow only one speaker at a time.

\noindent \textbf{How do specific expressions change between VC and F2F?} 
We have been looking at how the overall behavior changes between the two domains.
However, the key ability of our model is that we can delve deeper -- how do specific spatio-temporal patterns change across the two domains?
We aim to answer questions such as `How does a person's laugh change between the domains?'.
To answer this we extract spatio-temporal chunks with similar expressions.
We posit that since our partitioning function ($G_t$) segments clips into meaningful chunks and our translation predictor $G_f$ predicts similar~\translators~for similar chunks, clustering the parameters of the \translator~for a chunk should give us such meaningful spatio-temporal clusters. 

Therefore, we concatenate the $\omega$ and $\tau$ parameters for each chunk in our dataset and cluster them using k-means. 
We use the BIC-score~\cite{schwarz1978estimating} to find the optimal number of clusters. 
Many clusters are indeed meaningful. 
\cref{fig:main} (left) shows 2 examples each from 4 such clusters, with expressions such as ``speaking with raised eyebrows'' or ``head shake''.
We translate all VC chunks from a cluster to F2F and observe the distribution of latents specific to the cluster.
This analysis also results in many key insights. (see supplementary for more such insights)
\begin{figure}[t!]
  \centering
  \includegraphics[width=1.\linewidth]{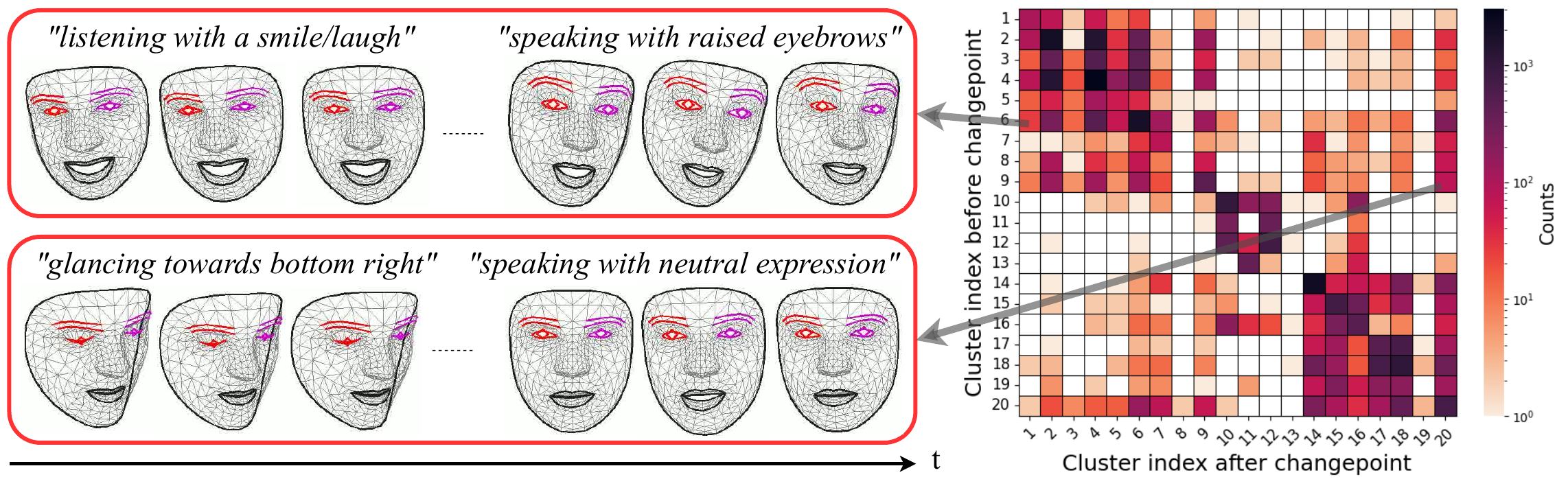}
  \vspace{-17pt}
  \caption{
    \textbf{\textit{Translator} cluster analysis.} We perform k-means clustering on all predicted~\translators~of the dataset and perform change-point analysis using them. \textbf{Right:} An entry $[i, j]$ in the matrix shows the frequency of timestamp changes while going from cluster $j$ to $i$. \textbf{Left: } We show two examples of frequently occurring cluster changes from \#1 to \#6 (\textbf{top}) and \#20 to \#9 (\textbf{bottom}). Note that the matrix would have been a diagonal if changepoint ($\tau$) prediction were not needed.
  }
  \label{fig:timestamp}
  \vspace{-15pt}
\end{figure}

\begin{itemize}[nosep,leftmargin=*]
    \item \textbf{Speakers tend to laugh bigger in F2F:} This can be observed with the ``Laugh'' latent code \#8 in the first ``speaking with a laugh'' cluster.
    Even though the peaks of the histogram do not shift by much, we can clearly see a distribution shift towards a bigger laughter (to the left).
    Note that a ``bigger speaker laugh in F2F'' is a different observation than ``more laughter in F2F'' that we discussed in the previous section.
    We could not have made this observation without the analysis of clusters shown in Fig. \ref{fig:main}
    \item \textbf{F2F Listeners do less head shakes:} This aligns with the previously reported results -- people emote more in VC~\cite{faith2023zoomgesture}.
    This is evident both from the translated faces and the distribution shift of the ``Head read steer'' 
    code (\#9).
\end{itemize}

\begin{figure}[tb]
  \centering
  \includegraphics[width=1.\linewidth]{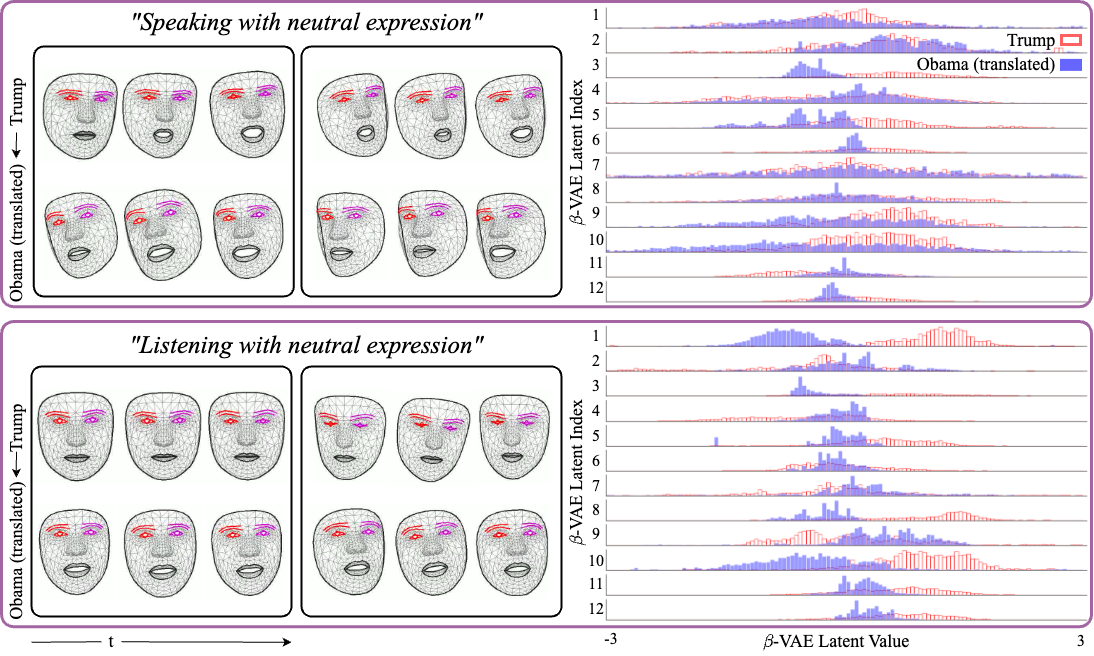}
  \vspace{-10pt}
  \caption{\textbf{Analysis of Presidents}: 2 \translator~clusters (Left) and the corresponding \facet~reports on the president (Obama and Trump) face dataset.}
  \label{fig:pres}
  \vspace{-5pt}
\end{figure}

\noindent \textbf{How do expressions change temporally?}
Another important aspect is how people transition from one expression to the next.
Since we are interested in temporally finer expression changes, we use the model with 7 \textbf{fixed-size chunks} and obtain \translators~for individual chunks.
We cluster these \translators~into 20 clusters and visualize common expression transitions.
We measure the number of times a transition happens from one cluster to another in consecutive chunks. 
\cref{fig:timestamp} (right) shows the number of such transitions from one cluster to another.

Using this we can infer the common expression transitions. 
We show examples of two frequently occurring transitions in \cref{fig:timestamp} (left).
The first transition is people switching from listening to speaking while keeping happy expressions.
Similarly, people frequently switch from listening and looking downwards to speaking neutrally.
Since a lot of values are off the diagonal, this also shows that our model can detect the changes due to the partitioning function $G_t$.

\noindent \textbf{Is \facet~generalizable to other tasks?} 
We also apply \facet~to a different task -- finding the difference between the facial expressions of two subjects, which in our case are presidents Obama (O) and Trump (T). Since this dataset is smaller, we initialize the training weights of the $\beta$-VAE using the ZoomIn $\beta$-VAE. A favorable side-effect is that the meanings of the latent codes do not change between the two datasets. 
However, there are some small differences due to the dataset distribution (refer to supplementary). 
\cref{fig:pres} shows histograms of distributions of latent codes when translating from Trump to Obama, for two clusters.
Several observations can be made from this.
\begin{itemize}[nosep,leftmargin=*]
\item \textbf{Trump has a more circular mouth when speaking.} This is evident from the clips from the speaking clusters as well as the left shift of the latent code \#6 distribution from Trump to Obama.
Note that this shift is not observable in the listening cluster.
Moreover, the distribution of latent code \#6 is spread out for Trump in both clusters indicating a wider range of horizontal mouth motion.
Both these observations have also been reported in the news~\cite{denby2015trumpface, collett2017trumpface}.
\item \textbf{Trump raises eyebrows more while listening.} Looking at the latent code \#11 we observe that Trump's eyebrows are more raised when listening in comparison to Obama and also in comparison to himself while speaking~\cite{golshan2017trumpface}.
\end{itemize}

\begin{figure}[t!]
  \centering
  \includegraphics[width=\linewidth]{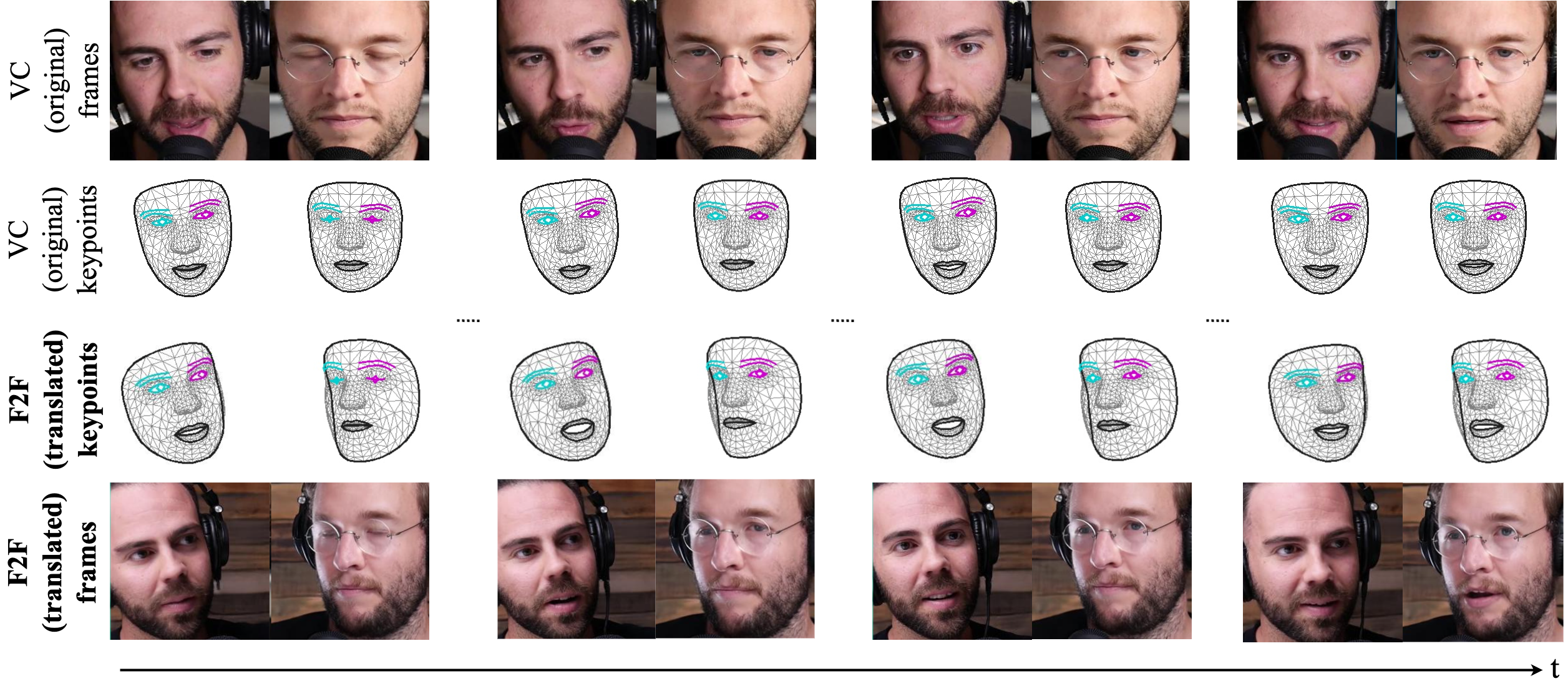}
  \vspace{-17pt}
  \caption{\textbf{De-zoomification} Given the facial keypoints of two people in a VC video,~\facet~generates the corresponding translated keypoints as if it was a F2F conversation, which are then converted to pixel-space (Sec.~\ref{ssec:dezoom}).
  }
  \label{fig:dezoom}
  \vspace{-15pt}
\end{figure}

\subsection{Applications (de-zoomification)}
We showed in our results that VC $\rightarrow$ F2F is more than just changing the eye gaze or head pose. 
We call this process \textit{de-zoomification}, and show that our method (explained in Sec.~\ref{ssec:dezoom}) can be used to convert an actual conversation that took place in a VC setting, into a F2F conversation as shown in \cref{fig:dezoom}. 
We observe that the eye blinks (first frame) and smiles (last frame) in \textit{de-zoomed} videos are consistent with the original.
The subjects also look more toward each other in the translated video which makes it feel like a F2F conversation.
This is an exciting application towards making virtual conversations feel more life-like. Please refer to the supplementary for videos.

\section{Discussion \& Conclusion}

\noindent\textbf{Limitations \& Future Work.} While our method provides detailed and interpretable reports, there is still room for improvement. 
First, our method relies on $\beta$-VAE for disentangling, which can introduce some noise as true disentanglement without imposing priors is hard ~\cite{locatello2019challenging}.
However, our method allows easy swap of $\beta$-VAE with better future alternatives.
Second, we currently demonstrate results on human facial keypoints only. 
While our method is general to any keypoint data, it is non-trivial to extend our architecture to images. \\

\noindent\textbf{Conclusion.} We addressed the task of discovering spatio-temporal patterns from unpaired facial keypoint data
by presenting~\facet, a generative domain translation method that is interpretable-by-design.
\facet~pre-computes domain agnostic, disentangled features, and learns interpretable linear transformations between domains. 
\facet~can predict temporal change-points that decouple expression changes from time.
Extensive experiments showcase~\facet's ability to generate informative reports of input-specific spatio-temporal patterns between domains. Lastly, we apply these learned differences to \textit{de-zoom} VC videos. \\

\noindent\textbf{Acknowledgements.} This research is based on work partially supported by the DARPA CCU program under contract HR001122C0034 and the National Science Foundation AI Institute for Artificial and Natural Intelligence (ARNI). PT is supported by the Apple PhD fellowship.

\bibliographystyle{ieeetr}
\bibliography{main}

\appendix

\renewcommand\thesection{\Alph{section}}
\renewcommand\thesubsection{\thesection.\arabic{subsection}}

\title{Supplementary Material for ``How Video Meetings Change Your Expression''} 

\author{}
\institute{}

\maketitle

\setcounter{figure}{8}
\setcounter{table}{1}

Video visualizations of qualitative results can be found at the following link: \href{https://facet.cs.columbia.edu/}{facet.cs.columbia.edu}
The supplementary material is structured as follows: in \cref{sec:s-dataset} we provide dataset preparation details, and dataset statistics. In \cref{sec:s-metrics} we provide additional discussion on the metrics used in our paper for evaluating our models. In \cref{sec:s-baselines} we provide a note on the baselines. In \cref{sec:s-implementation_details} we provide further implementation details which we hope can be useful for reproducing the results, and also provide more information on how the fixed translator-set baseline was implemented. In \cref{sec:s-ablations} we provide some more ablation results. Finally, in~\cref{sec:s-zoom_observations}, we provide additional insights about the differences between F2F and VC through~\facet, and in~\cref{sec:s-president_dataset}, we provide latent distributions and dataset-level statistics for the presidents data.

\section{Dataset details} \label{sec:s-dataset}

\begin{table}[tb]
  \caption{Key statistics about the ZoomIn dataset.
  }
  \label{tab:s-dataset}
  \centering
  \begin{tabular}{@{}c@{\hspace{0.5cm}}c@{\hspace{0.5cm}}c@{\hspace{0.5cm}}c@{}}
    \toprule
    \textbf{Mode of} & \textbf{Number} & \textbf{Average length of} & \textbf{Total duration of} \\
    \textbf{conversation} & \textbf{of videos} & \textbf{videos (minutes)} & \textbf{videos (hours)} \\
    \midrule
    Video conferencing  & 89  & 38.14 & 56.58  \\
    Face to face        & 279 & 39.44 & 184.70 \\
    \hline & \\[-2.0ex]
    Total               & 368 & 39.13 & 241.28 \\
  \bottomrule
  \end{tabular}
\end{table}

To study the differences between VC and F2F conversations, we collect the ZoomIn dataset, containing 240 hours of conversations in both VC and F2F settings.
We obtain these videos from the YouTube channel ``The Daily Talk Show - Australian Podcast'', where the hosts (mostly two specific individuals) have conversations on a wide range of topics.
Using background and eye gaze as cues, we are able to annotate the mode of conversation. We then run MediaPipe face landmark detector~\cite{lugaresi2019mediapipe} on all the frames to obtain 3D facial keypoints. 
Some examples can be found in \cref{fig:teaser}. 
Each sample clip in our dataset consists of detections on contiguous 176 frames from both the participants. In cases where the frame rate is higher, we apply linear interpolation on the keypoints to bring it down to 25 FPS. For the purpose of our experiments, we only consider those conversations where there are only two people.

Most of the videos in the ZoomIn dataset are of 1920 x 1080 resolution. We show some important statistics about our ZoomIn dataset in~\cref{tab:s-dataset}.
We highlight two positive aspects about this data. First, the conversations that we call as video conferencing (VC) do not suffer from any quality issues of the second party. This is because both the participants record their videos on their own systems, and then collate it together for best visual quality for their viewers. Second, in all the conversations, the cameras remains fixed. Hence both the participants have a camera in front of them at most times, and we can view the facial expressions of both the participants simultaneously during the course of the conversation. 

Additionally, to study differences in communication styles across subjects, we collect a second dataset of presidential speaking styles. Specifically, we collect videos of statements and announcements by U.S. presidents Obama and Trump. We would like to highlight that the presidents dataset is much smaller (around 10$\times$) as compared to the ZoomIn dataset, and suffers from issues like inconsistent and moving camera positions. Also, we perform undersampling to overcome the huge class imbalance in the presidential dataset.

To the best of our knowledge, this is a \textit{first-of-its-kind} dataset that contains information about the variation of the mode of conversation between VC and F2F conversations. Please refer to \href{https://facet.cs.columbia.edu/}{facet.cs.columbia.edu} for more on data.

\section{Metrics} \label{sec:s-metrics}

Evaluating discovered differences is extremely challenging in settings without supervision as we cannot evaluate what will be discovered beforehand.
Additionally, to the best of our knowledge, there are no existing prior works that tackle this problem of discovering differences from spatio-temporal keypoint data. Existing generative research (e.g., GANs) uses human evaluation or inception score.
Firstly, recall that our task is to discover differences \textit{without} access to labels for such differences. 
Since humans also do not have prior expert knowledge, we cannot run human studies. Secondly, our dataset (containing only 2 classes) has a few dominant patterns distinguishing them (e.g., `Head Pitch'). As we show in \cref{fig:bar}, training a classifier directly results in good accuracy by only picking up on a few dominant features. Thus, getting a high or low inception score is also not a useful metric.

Therefore we evaluate how various design choices of our interpretable-by-design model result in finding good differences.
The key metric we use to evaluate this is how well the translation function $G_{XY}$ fools the discriminator while being interpretable. 
So we measure the the discriminator accuracy in distinguishing real samples in a domain versus the samples translated to that domain.
A model that leads to \emph{lower discriminator accuracy} can capture more modes of variations and is therefore a better translator. 
We hypothesize that in the long term for a discriminator to perform well, it needs to identify differences that our constrained generator cannot produce, but can be used by the discriminator for detecting fake samples. Those differences that the generator can learn to rectify, reach a state of equilibrium with the discriminator. Hence we find this metric to stabilize after a certain set of epochs, and therefore we provide the average over the last 100 epochs.

\begin{table}[ht!]
  \caption{Statistical analysis for \cref{tab:ablation} (for $c=2$ on ZoomIn dataset) with 95\% confidence intervals from 8 runs.}
  \centering
  \begin{tabular}{@{}ll@{\hspace{0.1cm}}ccc@{\hspace{0.1cm}}ccc@{}}
    \toprule
    \multicolumn{2}{c}{\textbf{Model Type}}  & \multicolumn{3}{c}{\textbf{Disc. Acc. (\%) $\downarrow$}} \\
    \cmidrule(r{1em}){1-2}
    \cmidrule(lr{2em}){3-5}
        $G_f$ & $G_t$ (chunks) & Avg & \hspace{0.cm}F2F$\rightarrow$VC & \hspace{0.cm}VC$\rightarrow$F2F  \\
    \midrule
    \multirow{2.5}{0.08\linewidth}{Fixed}
    & Fixed-size & 94.24 $\pm$ 0.84 & 94.94 $\pm$ 0.74 & 93.57 $\pm$ 1.54 \\
    \cmidrule{2-5}
    & Var. & 92.51 $\pm$ 0.30 & 93.63 $\pm$ 0.63 & 91.50 $\pm$ 0.77 \\
    \midrule
    \multirow{2.5}{0.08\linewidth}{Pred.}
     &Fixed-size & 80.81 $\pm$ 1.33 & 80.38 $\pm$ 2.61 & 81.10 $\pm$ 2.64 \\
    \cmidrule{2-5}
    & {Var. (\textbf{\facet})} & \textbf{74.11 $\pm$ 0.89} & \textbf{74.70 $\pm$ 1.81} & \textbf{74.26 $\pm$ 1.01} \\
    \bottomrule
  \end{tabular}
  \label{tab:stats}
\end{table}

We find the accuracy of the discriminator trained at the \textit{original} task of detecting video-calls (VC) vs face-to-face (F2F) conversations to be 99\%, showing that the discriminator is strong and able to perform this task well.
The discriminator accuracy of 73.16\% (\cref{tab:ablation}) shows that our method is able to fool the discriminator and that our method is able to detect and remove the differences between the domains.
We also provide statistical analysis for \cref{tab:ablation} in \cref{tab:stats}.

\section{Baselines} \label{sec:s-baselines}

Note that \textbf{Fixed translator-set} is a variation on $G_f$, whereas \textbf{No Partitions} and \textbf{Fixed translator-set} are variations on $G_t$.
Therefore we can mix and match them for evaluation as well.

\section{Implementation Details}\label{sec:s-implementation_details}
Our input data consists of by 478 3D facial keypoints.
For the spatial $\beta$-VAE, we use a 5-layer MLP (with a LeakyReLU non-linearity) as the encoder as well as the decoder (with layer dimensions as 512, 512, 256, 256, 128; similarly in the reverse direction for the decoder). 
We begin training from solely using the training reconstruction loss (i.e. $\beta=0$) and gradually increase it. We choose the highest $\beta$ that does not compromise on reconstruction.
We train the $\beta$-VAE with 16 latent codes but find that only 12 of them show any variation ($l=12$), capturing various dimensions of expressions that can be seen in Fig. 1. We measured this by varying a latent between ($-3, +3$), and measuring the mean L2 distance between reconstruction keypoints. For a $\beta$-VAE (with $l$$=$16), we observe that the 12 useful dimensions have a variation at least \emph{138 times} the remaining 4. Lastly, we use Adam optimizer with a learning rate of 0.001.

For the translation function, we use 3-layer MLP (LeakyReLU non-linearity) for $G_f$ and $G_t$. The discriminator is also a 3-layer MLP (dropout of 0.5). We use a factor of 8 to decrease the dimensionality in the subsequent layers as needed.
We use Adam optimizer with a learning rate of $10^{-4}$. 
We train for 1000 epochs on \textit{ZoomIn} and for 2000 epochs on the presidents dataset.
Our data is split in 90:10 train-test ratio. 
We train using an Nvidia RTX 2080Ti. 
For the translator function $G_t$, we explored different values of the temperature $Q$, with no impact 
on performance. We set $Q=0.12$ for our experiments. For training vid2vid (domain transfer), we use 8 Nvidia A6000 GPUs, and train for 50 epochs. 

\subsection{Fixed translator-set baseline implementation details}
Instead of learning to predict a~\translator~from the input, the fixed translator-set baseline instead chooses from a list of possible~\translator s.
As stated in the main paper, this alternative model learns two things.
First, is learns a classifier that chooses one out of $p$ choices.
Second, it also simultaneously learns a list of parameters to choose from $P\in\mathbb{R}^{p\times 2l}$.

Even in this case, the model is non-differentiable as the \emph{choosing} operation using the output of the classifier is non-differentiable.
However, we can approximate the $\argmax$ operation in the classifier with a softmax to make it differentiable.
An additional issue with this approach is that, since there is no explicit supervision for the model to always predict a single choice out of $p$, the model cannot learn to predict a single high-confidence choice.
Therefore, we additionally use an entropy loss term to enforce the model to produce \emph{peaky} or high-confidence choices.
If $s\in\mathbb{R}^p$ is the softmax probability score for each choice, the entropy loss can be written as:
$$\mathcal{L}_{\text{entropy} = -\sum_{i=1}^{p} s_i \log (s_i)}$$

The predicted parameter values can be written as the weighted sum of the $p$ choices.
$$\omega^* = \sum_{i=1}^{p} s_i\omega_i \qquad\qquad \phi^* = \phi_{i=1}^{p} s_i\phi_i$$

Where, $\omega_i$ and $\phi_i$ are the $i^{\text{th}}$ choice for multiplicative and additive factors.
As the training progresses, the entropy loss forces the model to select only one out of $p$ choices.
The translation function $G_{XY}$ can be be written as,

$$G_{XY} (z_x) = \omega^*\odot z_x + \phi^* $$

The entropy loss term is used along with the main adversarial loss term leading to a solution for the classifier and the learned list of parameters. In our experiments, we used $p=32$.

\section{Ablations on the~\translator} 
\label{sec:s-ablations}

\begin{table}[ht!]
  \caption{Ablations over shifting and scaling in the~\translator.}
  \label{tab:s-ablation}

  \centering
  \begin{tabular}{@{}llr@{\hspace{0.3cm}}ccc@{\hspace{0.3cm}}ccc@{}}
    \toprule
    \multicolumn{3}{c}{\textbf{Model Type}}  & \multicolumn{6}{c}{\textbf{Discriminator Acc. (\%) $\downarrow$}} \\
    & &  & \multicolumn{3}{c}{ZoomIn} & \multicolumn{3}{c}{Presidents} \\
    \cmidrule(r{1em}){1-3}
    \cmidrule(lr{2em}){4-6}
    \cmidrule(lr{0.0em}){7-9}
        $G_f$ & $G_t$ & c & Avg & \hspace{0.cm}F2F$\rightarrow$VC & \hspace{0.cm}VC$\rightarrow$F2F  & Avg & \hspace{0.1cm}O$\rightarrow$T & \hspace{0.1cm}T$\rightarrow$O  \\
    \midrule
    \multirow{6}{0.12\linewidth}{Predicted translator} 
     &Var. Chunks ($\times$) & 2 & 86.50 & 86.71 & 86.31 & 92.03 & 88.52 & 95.33 \\
     &Var. Chunks ($\times$) & 7 & 86.29 & 86.69 & 85.98 & 91.82 & 89.86 & 93.44 \\
     \cmidrule{2-9}
     &Var. Chunks (+) & 2 & 81.72 & 81.91 & 81.56 & 96.52 & 96.22 & 97.05 \\
     &Var. Chunks (+) & 7 & 81.70 & 82.16 & 81.34 & 96.51 & 96.14 & 97.08 \\
     \cmidrule{2-9}
    & {Var. Chunks (\textbf{\facet})} & 2 & 73.28 & 73.33 & \textbf{73.50} & 79.35 & 70.31 & \textbf{88.70} \\
    & {Var. Chunks (\textbf{\facet})} & 7 & \textbf{73.16} & \textbf{72.65} & 73.92 & \textbf{78.14} & \textbf{67.50} & 89.06 \\
    \bottomrule
  \end{tabular}
\end{table}

To study the importance of each component of our~\translator, we perform an ablation by removing each component systematically. The results are shown in~\cref{tab:s-ablation}. In rows 1 \& 2, the~\translator~predicts the scaling factor $\omega$ only (no shift), denoted by Var. Chunks ($\times$). In rows 3 \& 4, the~\translator~predicts the shift $\phi$ only (no scaling), denoted by Var. Chunks (+).

For the ZoomIn dataset, comparing Var. Chunks ($\times$) with Var. Chunks (+), we observe that shift ($\phi$) is consistently more important than scale ($\omega$), as removing it results in worse performance than removing scale. However, this trend is reversed for the Presidents dataset, where scale ($\omega$) is more important than shift ($\phi$).
\facet~consistently outperforms both these ablations across both datasets, highlighting the need for predicting both shift and scale to ensure good domain transfer.

\section{ZoomIn Observations} \label{sec:s-zoom_observations}
We present additional insights that~\facet~can discover about the differences in face-to-face and video conversations. Please refer to \cref{fig:main} and \cref{fig:results-faces}.

\noindent \textbf{People move their head more during a F2F conversation.} 
See \cref{fig:main} first two rows (or clusters).
From the latent code \#9 and \#10 in both clusters (speaking with a laugh; listening with neutral expression) we notice that people move their heads a lot more in an F2F conversation.
Note that this is different from saying ``People can be facing at different points with respect to the camera in an F2F conversation''. 
The latter observation can be made by the bimodal F2F distribution of latent \#9 and \#10.
The F2F distribution is more spread out than the VC distribution.
This suggests that not only are the views shifted w.r.t. the camera in F2F conversation, but the head also moves more.

\noindent \textbf{People open their mouth more while listening in a F2F conversation.} 
See \cref{fig:main} row 2 (cluster: ``listening with a neutral expression''). This can be inferred by looking at latent \#3 as well as the translated conversation where the original VC shows pursed lips while the translated F2F conversations shows slightly more opened lips. This is consistent with observations in neuroscience which show that people are more self-conscious due to the presence of a self-viewing screen in video chats, leading to increased self-regulating of facial expressions~\cite{ShinEffects2022}.

\noindent \textbf{Eyebrow raises are bigger in VC.}
See \cref{fig:results-faces}.
This again follows through the fact that ``people emote bigger during in a VC''~\cite{faith2023zoomgesture}.
In the speaking with a raised eyebrow cluster, we observe that the VC distribution shifted towards the right (bigger eyebrow raise) for latent \#11.
This suggests that even though people raise eyebrows in both modes of conversation, the magnitude of the raise is larger during a VC conversation.

\begin{figure}[ht!]
  \centering
  \includegraphics[width=0.8\linewidth]{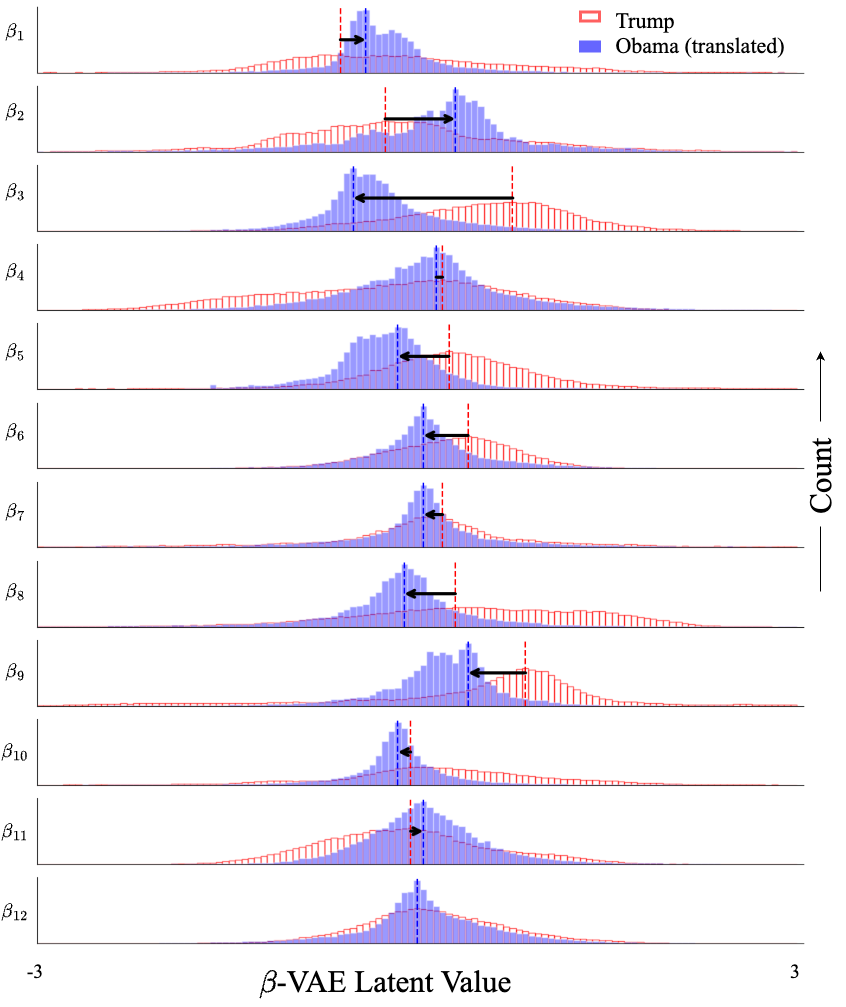}
  \caption{
    \textbf{Latents for presidents data.} We vary each dimension of the 12-dimensional latent obtained through $\beta$-VAE encoding by keeping other dimensions fixed (rows). We show the dataset-level statistics for the desired latent, while the dotted line shows the modes. The direction and length of the arrows show the extent to which different latents change across domains on average. 
    Please refer to \href{https://facet.cs.columbia.edu/}{facet.cs.columbia.edu} to visualize the latents. This is analogous to \cref{fig:results-faces} (which was for ZoomIn dataset).
  }
  \label{fig:results-presidents}
\end{figure}

\noindent \textbf{Eyes are more closed in VC conversation.}
See \cref{fig:results-faces}.
Although the latent codes \#3 and \#12 are not fully disentangled, we observe that going right on both corresponds to closing eyes.
We observe that people close their eyes more during a VC as the F2F distribution shifts to the left.
This can be partially attributed to zoom fatigue~\cite{nesher-22,Fauville_zoom2021}, however, a deeper understanding is needed to better understand this. We cannot examine whether this is due to more blinking or just more closed eyes.
Our model lacks a disentangled latent for eyelid and eye movement, making it harder to delve deeper. 
We believe that better disentanglement (e.g., by using eye-specific loss-terms) and better keypoint tracking can lead to a more in-depth examination of this observation.

\section{Presidents Dataset Stats} \label{sec:s-president_dataset}
Similar to \cref{fig:results-faces} where we provided dataset-level statistics for the ZoomIn dataset, we also provide them for the Presidents data for completeness in \cref{fig:results-presidents}.

\end{document}